\documentclass[runningheads]{llncs}
\usepackage{makecell}
\usepackage{graphicx}
\usepackage{hyperref}
\usepackage{float}
\usepackage{siunitx}
\usepackage{amssymb}
\usepackage{amsmath}
\usepackage{bbding}
\usepackage[marginal]{footmisc}

\begin{document}
\title{CTooth: A  Fully Annotated 3D Dataset and Benchmark for Tooth Volume Segmentation on Cone Beam Computed Tomography Images}
%
\titlerunning{CTooth: A Fully Annotated 3D Dataset and Benchmark}
%
%
\author{Weiwei Cui\inst{2}, Yaqi Wang \inst{3}\inst{(}\Envelope\inst{)}, Qianni Zhang \inst{2}, Huiyu Zhou \inst{4},  Dan Song \inst{5}, Xingyong Zuo \inst{5}, Gangyong Jia \inst{1}, {Liaoyuan Zeng  \inst{5}}\inst{(}\Envelope\inst{)}}

\authorrunning{Weiwei et al.}
%
\institute{Hangzhou Dianzi University\\ 
\and
Queen Mary University of London \\
\and
Communication University of Zhejiang\\
\and
University of Leicester\\
\and 
University of Electronic Science and Technology of China}

%
\maketitle              

\begin{abstract}
3D tooth segmentation is a prerequisite for computer-aided dental diagnosis and treatment. However, segmenting all tooth regions manually is subjective and time-consuming. Recently, deep learning-based segmentation methods produce convincing results and reduce manual annotation efforts, but it  requires a large quantity of ground truth for training. To our knowledge, there are few tooth data available for the 3D segmentation study. In this paper, we establish a fully annotated cone beam computed tomography dataset CTooth with tooth gold standard. This dataset contains 22 volumes (7363 slices) with fine tooth labels annotated by experienced radiographic interpreters. To ensure a relative even data sampling distribution,  data variance is included in  the CTooth including missing teeth and dental restoration. Several state-of-the-art segmentation methods are evaluated on this dataset.  Afterwards, we further summarise and apply a series of 3D attention-based Unet variants for segmenting tooth volumes. This work provides a new benchmark for the tooth volume segmentation task. Experimental evidence proves that attention modules of the 3D UNet structure  boost responses in tooth areas and inhibit the influence of background and noise. The best performance is achieved by 3D Unet with SKNet attention module, of 88.04 \% Dice and 78.71 \% IOU, respectively. The attention-based Unet framework outperforms other state-of-the-art methods on the CTooth dataset. The codebase and dataset are released  \href{https://github.com/liangjiubujiu/CTooth}{here}.

\keywords{3D dental dataset  \and Tooth segmentation  \and Attention.}
\end{abstract}
\section{Introduction}
Due to the low dose and high-definition vision, dental cone beam computed tomography (CBCT) film  has been a standard to exam teeth conditions during the preoperative stage. 3D tooth segmentation for CBCT images is a prerequisite for the orthodontics surgery since it assists to reconstruct  accurate 3D tooth models. However, manual tooth annotations requires domain knowledge of dental experts and it is time and labour-consuming. Therefore, automatic and precise tooth volume segmentation is essential.

Some shallow learning-based methods try to quickly segment teeth from X-ray or CBCT images such as region-based \cite{lurie2012recursive}, threshold-based \cite{ajaz2013dental},boundary-based \cite{hasan2016automatic} and cluster-based \cite{alsmadi2018hybrid} approaches. However, these methods present limited evaluation results on small or private datasets, and segmentation performances are highly dependent on the manual feature engineering. The main reason causing these
problems is the lack of an open  dental datasets with professional annotations.  Dental X-ray Image  dataset   is the first public dental dataset published on the ISBI Grand Challenges 2015 mainly focusing on dental landmark detection and caries segmentation \cite{wang2016benchmark}. LNDb  consists of  annotated dental panoramic X-ray images. All teeth in patients' oral cavities are identified and marked  \cite{silva2018automatic}. Teeth\_dataset  is a relative small classification dataset, which presents single crowns and is labelled with or without caries \cite{A:2005}. We review two widely-used dental X ray datasets. However, the existing dental dataset all record 2D dental images and the spatial teeth information is compressed and distorted.

Recently, deep learning-based methods are
available with an attempt to solve 3D tooth segmentation. Some methods apply Mask R-CNN  on tooth segmentation and detection \cite{Jader2018DeepIS,cui2019toothnet}. These approaches require accurate and specialised instance dental labels which are not always available. Other methods exploit 3D Unet  with well-designed initial dental masks or complex backbones \cite{wu2020center,yangdeep}. However, these segmentation methods are evaluated on private tooth datasets.

Attention mechanism has been widely used in computer vision tasks  and achieve state-of-the-art performances on medical image segmentation tasks. To the best of our knowledge, attention based strategies \cite{woo2018cbam,fu2019dual,li2019selective} have not yet been applied in solving 3D tooth segmentation tasks on CBCT images mainly due to the annotated data limitation. Inspired by the success of attention mechanism on other medical image processing tasks, this paper's motivation is to explore the effectiveness of attention mechanisms in the tooth volume segmentation method.
 
 In this work,  we aim to (1) build the first open-source 3D CBCT dataset CTooth with dental annotations. Several state-of-the-art segmentation methods are evaluated on the CTooth. (2) propose an attention-based segmentation framework as a new benchmark. Experiment results show that the attention-based framework outperforms other 3D segmentation methods in this task.

\section{CTooth dataset: Fully Annotated Cone Beam Computed Tomography Tooth Dataset }

\subsection{Dataset Summary}
Several 2D open-access dental datasets have been proposed, as shown  in Tab. \ref{tab.1}. Dental X-ray Image dataset  contains 400 lateral cephalometry films and 120 bitewing images, and seven types of areas are marked, including caries, crown, enamel, dentin, pulp, root canal treatment and restoration \cite{wang2016benchmark} .  LNDb    consists of 1500 panoramic X-ray images categorized by 10 classes, with a resolution of 1991 by 1127 pixels for each image \cite{silva2018automatic}. It is suitable for 2D tooth detection and semantic segmentation study. A small caries classification dataset Teeth\_dataset with 77 intraoral images is proposed on Kaggle \cite{A:2005}. A shallow VGG network  has easily handled the binary caries classification task on this dataset \cite{simonyan2014very}. These dataset have limit quantity of images and not suitable for segmenting tooth volume by deep learning based method. Therefore, our first contribution is to build a 3D CBCT dental dataset with teeth annotations.

 \begin{figure}[t]
\centering
	\includegraphics[width=\textwidth]{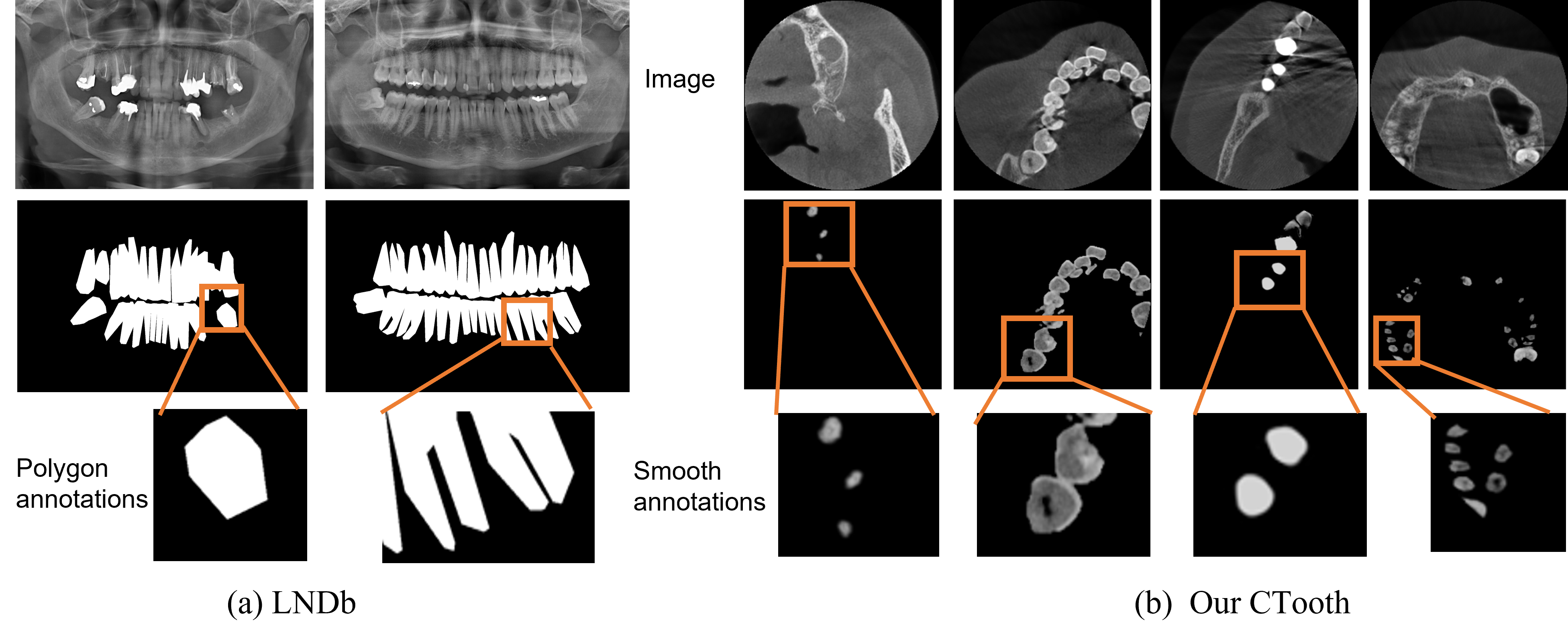}
	\caption{Data comparison between the 2D LNDb dataset and our 3D Ctooth dataset. }
	\label{fig.4}
\end{figure}

\begin{table}[b]
\caption{Summary of publicly available dental datasets. To our knowledge, there are no 3D public dataset with tooth volume annotations.}
\setlength{\tabcolsep}{1mm}{
\begin{tabular}{|l|l|l|l|l|}
\hline
Dataset                                                               & Year & modality       & task                                                                               & scans                                             \\ \hline
\begin{tabular}[c]{@{}l@{}}Dental X-ray\\ Image  \cite{wang2016benchmark} \end{tabular}  & 2015 & 2D X ray       & \begin{tabular}[c]{@{}l@{}}Caries segmentation  \\ Landmark detection\end{tabular} & \begin{tabular}[c]{@{}l@{}}120\\ 400\end{tabular} \\ \hline
LNDb \cite{silva2018automatic}                                                                 & 2016 & 2D X ray       & Tooth segmentation and detection                                                   & 1500                                              \\ \hline
Teeth\_dataset \cite{A:2005}                                                        & 2020 & 2D RGB image   & Caries classification                                                                 & 77                                                \\ \hline
Our Ctooth                                                            & 2022 & 3D CBCT  & Tooth volume segmentation                                                          & 7363                                              \\ \hline
\end{tabular}}
\label{tab.1}
\end{table}

The images used in our data set were acquired on an OP300, manufactured by Instrumentarium Orthopantomograph$^{\circledR}$, and acquired at the University of Electronic Science and Technology of China Hospital. The radiographic images used for this research were coded in order to avoid identifying the patients.  All CBCT slices originally obtained by the orthopantomography have  266 $\times$ 266 pixels in the axial view.  The in-plane resolution is about  $0.25  \times  0.25 mm^{2}$ and the slice thickness range from 0.25mm to 0.3 mm.  The gathered data set consists of 5504 annotated CBCT images from 22 patients. There are in total  5803 CBCT slices, out of which 4243  contain tooth annotations. The images contain significant structural variations in relation to the teeth position, the number of teeth, restorations, implants, appliances,  and the size of jaws.

\subsection{Expert Annotation}

Figure \ref{fig.4} illustrates a few samples of  the LNDb dataset \cite{silva2018automatic}. This is a 2D tooth dataset based on the panoramic X-ray images. These X-ray scans were first transformed to grey-scale images, causing a compressed range of pixel values. In addition, experts marked all tooth regions  using the polygon annotations on the transformed X-ray images.  The inaccurate polygon mode leads to boundary error during model training.

Different from the LNDb, all scans in the CTooth dataset are annotated in a smooth mode. Also, we only resize all CBCT slices to 256×256 resolution  and do not process any transformation to change CT values. Four trainees from a dental association (with four years of experience)  manually mark all teeth regions.  They first use ITKSNAP \cite{yushkevich2006user} to delineate tooth regions slice-by-slice in axial view. Then the annotations are fine-tuned manually in the coronal view and sagittal view. In the annotation stage for each volume, it roughly takes 6 hours to annotate all tooth regions and further requires 1 hour to check and refine the annotations. The CTooth dataset took  us around 10 months to collect, annotate and
review. With its data amount and quality, we believe it is a valuable and desired asset to share in public for computer-aided tooth image research.

\begin{table}
\centering
\caption{Categories in the UCTooth dataset, the number of patients in each category, average number of teeth and slices per patient and the average statics of Hounsfield pixel values.}
\begin{tabular}{|c|c|c|c|c|}
\hline
Category& \# Volume&  \# Ave teeth& \# Ave slices& Mean CT value\\
\hline
Missing teeth w appliance	&6&13&217&140\\
\hline
Missing teeth w/o appliance&5&10&260&156 \\
\hline

Teeth w  appliance &6&16&351&144\\
\hline

Teeth w/o  appliance&5&12&220&153 \\

\hline
\end{tabular}

\label{tab.2}
\end{table}

According to the features of the annotated tooth areas, all the CBCT volumes are categorized in to four different classes  in terms of missing teeth and dental appliance.  Table \ref{tab.2} summarises the basic statistics including average number of volumes in each category, average number of teeth, average slices per volume, and the mean pixel values (Hounsfield) for each category.   According to the statistical results, we observe that the data distribution is balanced among four categories.

\section{Attention-based Tooth Volume Segmentation}
\subsection{Framework Design}
\begin{figure*}[t]
\centering
	\includegraphics[width=\linewidth]{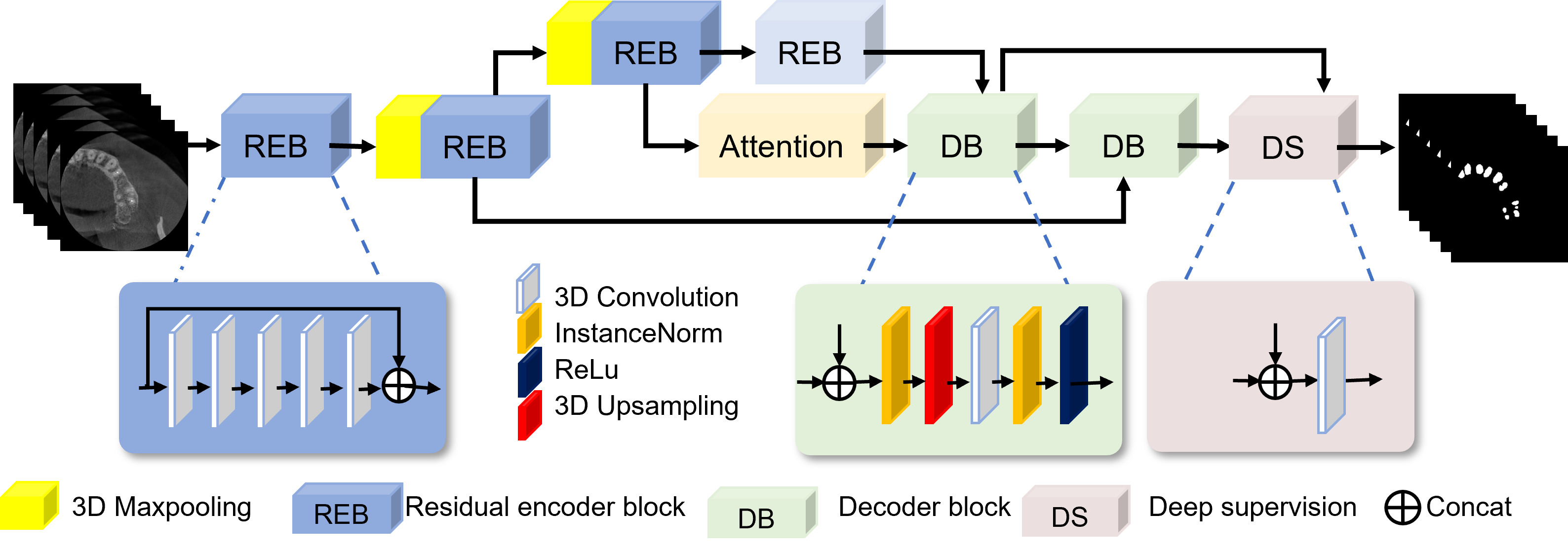}
	\caption{The architecture of the proposed BM-Unet.}
	\label{fig.5}
\end{figure*}

To segment all tooth regions precisely, especially in small tooth roots, we  propose an attention-based segmentation benchmark on the CTooth dataset as shown in Figure \ref{fig.5}. Inspired by the excellent performance of Unet based
networks for medical image segmentation and classification,
we extend the pipeline of Unet \cite{ronneberger2015u} to a 3D version
as our base framework. Then we attempt to insert an attention branch at the bottleneck position of the base model to boost the tooth segmentation performances.

In the backbone feature extraction module, a batch of CBCT sub-volumes first pass through
three Residual Encoder Blocks (REBs). Each REB contains 5 3D convolution layers with a shortcut connection, and is designed to collect tiny tooth features. The features from REB then go through a 3D maxpooling layer to increase the receptive field represented by each voxel. The encoded features with the largest receptive field are fed into an attention module and a  bottleneck module (REB), separately. After that, the encoded tooth features from the bottleneck and the attention modules  are reconstructed by a series of fully convolutional decoders (DB). The decoder design is the same as the decoder of 3D Unet including instanceNorm, 3D upsampling, 3D convolution and ReLu.   Since the teeth size is relatively similar but with  variation, we finally improve the response of the tooth areas with a deep supervision (DS) step. In the testing period, we attempt to explore the efficiency of various attention modules on our framework including channel attention, spatial attention and self attention.

\subsection{Loss Function}
Let $\mathbf{R}$ be the ground truth with voxel values $r_{n}$, and  $\mathbf{P}$ be the predicted probabilistic map with the tooth label over $N$ voxels $p_{n}$. The element-wise dice similarity coefficient (DSC) \cite{milletari2016v,sudre2017generalised} between $\mathbf{R}$ and $\mathbf{P}$  is  defined as follows:
\begin{equation}
    \mathrm{DSC}=\frac{2 \sum_{n=1}^{N} p_{n} r_{n}+\epsilon}{\sum_{n=1}^{N} p_{n}+r_{n}+\epsilon}+\frac{2  \sum_{n=1}^{N}\left(1-p_{n}\right)\left(1-r_{n}\right)+\epsilon}{\sum_{n=1}^{N} 2-p_{n}-r_{n}+\epsilon}.
\end{equation}

The $\epsilon$ term is used here to ensure the loss function's stability by avoiding the
numerical issue of dividing by 0.
To cope with class imbalance between tiny tooth roots and background, the loss function of model training is defined by:
\begin{equation}
    L=1-\frac{2 w_{1}  \sum_{n=1}^{N} p_{n} r_{n}+\epsilon}{\sum_{n=1}^{N} p_{n}+r_{n}+\epsilon}+\frac{2  w_{2} \sum_{n=1}^{N} \left(1-p_{n}\right)\left(1-r_{n}\right)+\epsilon}{\sum_{n=1}^{N} 2-p_{n}-r_{n}+\epsilon},
    \label{equ.12}
\end{equation}
where $w_{1}$ and $w_{2}$ are weights, and $w_{1}+w_{2}=1$. 
\section{Experiments and results}

\subsection{Evaluation Metrics on the CTooth}
 The segmentation inference results are evaluated  using  dice similarity coefficient (DSC), weighted dice similarity coefficient (WDSC), intersection over union (IoU), sensitivity (SEN), positive predictive value (PPV), Hausdorff  distance  (HD), average symmetric  surface  distance  (ASSD), surface overlap (SO) and surface dice (SD) \cite{heimann2009comparison} .

Two  distance metrics (HD and ASSD) are evaluated on the surfaces of tooth volumes of ${\mathbf{R}}$  Let $\mathbf{S}_{\mathbf{R}}$ be a set of surface voxels of  $\mathbf{R}$,  and the  shortest  distance  of  an  arbitrary  voxel $p$ with $\mathbf{S}_{\mathbf{R}}$ can  be defined as follows:
\begin{equation}
d\left(p, \mathbf{S}_{\mathbf{R}}\right)=\min _{s_{\mathbf{R}} \in \mathbf{S}_{\mathbf{R}}}\left\|p-s_{\mathbf{R}}\right\|_{2},
\end{equation}
where $s_{\mathbf{R}} \in \mathbf{S}_{\mathbf{R}}$ and $\Vert \Vert$ is the distance paradigm between point sets, e.g. Euclidean distance. Thus, HD is defined as follows:
\begin{equation}
HD=\max \left\{\max _{s_{\mathbf{R}} \in \mathbf{S}_{\mathbf{R}}} d\left(s_{\mathbf{R}}, \mathbf{S}_{\mathbf{P}}\right)+\max _{s_{\mathbf{P}} \in \mathbf{S}_{\mathbf{P}}} d\left(s_{\mathbf{P}}. \mathbf{S}_{\mathbf{R}}\right)\right\}.
\end{equation}

The distance function is defined as:
\begin{equation}
D\left(\mathbf{S}_{\mathbf{R}}, \mathbf{S}_{\mathbf{P}}\right)=\Sigma_{s_{\mathbf{R}} \in \mathbf{S}_{\mathbf{R}}} d\left(s_{\mathbf{R}}, \mathbf{S}_{\mathbf{P}}\right).
\end{equation}

Moreover, the ASSD can be defined as follows:
\begin{equation}
\operatorname{ASSD}(\mathbf{R}, \mathbf{P})=\frac{1}{\left|\mathbf{S}_{\mathbf{R}}\right|+\left|\mathbf{S}_{\mathbf{P}}\right|}\left(D\left(\mathbf{S}_{\mathbf{R}}, \mathbf{S}_{\mathbf{P}}\right)+D\left(\mathbf{S}_{\mathbf{P}}, \mathbf{S}_{\mathbf{R}}\right)\right),
\end{equation}
where $\left| \cdot \right|$ is the number of points of the set.

The surface overlapping  values $SO$ of $\mathbf{S}_{\mathbf{P}}$ is calculated by  
\begin{equation}
o(p)=\left\{
\begin{aligned}
1, \qquad d\left(p, \mathbf{S}_{\mathbf{R}}\right)<\theta \\
0, \qquad d\left(p, \mathbf{S}_{\mathbf{R}}\right)>\theta\\
\end{aligned}
\right.
\label{equ.17}
\end{equation}
\begin{equation}
SO(\mathbf{S}_{\mathbf{P}})=\frac{\sum_{n=1}^{M_{\mathbf{P}}} o(p_{n})}{M_{\mathbf{P}}},
\label{equ.18}    
\end{equation}
where $\theta$ is a maximal distance to determine whether two points have the same spatial positions. $M_{\mathbf{P}}$ is the number of points in the surface set $\mathbf{S}_{\mathbf{P}}$.

The surface overlapping dice values of $\mathbf{S}_{\mathbf{R}}$ and $\mathbf{S}_{\mathbf{P}}$ are calculated by:
\begin{equation}
    SD=DC(r_{m},p_{m}),
\end{equation}
where $o(r_{m})=1$  and $o(p_{m})=1$.

\subsection{Experimental results}
To reduce the noise and increase image contrast, we apply contrast limited adaptive histogram equalization \cite{pisano1998contrast} on image slices. The transformed input images are normalized to the range  [0,1]  for each voxel. Kaiming initialization \cite{he2015delving} is used for initializing all the weights of the proposed framework.  The Adam optimizer is used with a batch size of 4 and a learning rate of 0.0004 with a step learning scheduler (with step size=50 and $\gamma=0.9$ ). The learning rate is decayed by multiplying  0.1  for every 100 epochs. The weighted dice loss is applied to guide the network backpropagation \cite{ma2021loss} with $w_{1}=0.1$ and $w_{2}=0.9$. The network is trained for 600 epochs using an Intel(R) i7-7700K desktop system with a 4.2 GHz processor, 16GB  memory, and 2 Nvidia GTX 1080Ti  GPU machine.  We implement this framework using the Pytorch library \cite{paszke2019pytorch}. It takes  10 hours to complete all the training and inference procedures.

\begin{table}[h]
\centering
\caption{Evaluation comparison among differnet tooth volume segmentation methods on the CTooth dataset.}
\setlength{\tabcolsep}{3.5mm}{
\begin{tabular}{|c|c|c|c|c|c|}
\hline
Method&WDSC&DSC&IOU&SEN&PPV\\

\hline

DenseVoxelNet \cite{yu2017automatic}&79.92&57.61&49.12&89.61&51.25\\
\hline
3D HighResNet \cite{li2017compactness} &81.90&61.46&52.14&87.34&59.26\\
\hline
3D Unet \cite{cciccek20163d}&82.00&62.30&52.98&88.57&60.00\\
\hline
VNet \cite{milletari2016v}&82.80&63.43&55.51&87.47&64.64\\
\hline
\textbf{Ours}&\textbf{95.14}&\textbf{88.04}&\textbf{78.71}&\textbf{94.71}&\textbf{82.30}\\
\hline

\end{tabular}}

\label{tab.111}
\end{table}

\begin{table}[b]
\centering
\caption{Ablation study results of choosing various attention modules in the proposed attention-based framework.}
\setlength{\tabcolsep}{1.7mm}{
\begin{tabular}{|l|l|l|l|l|l|l|l|l|l|l|}
\hline
Attention  & DSC   & IOU   & SEN   & PPV   & HD    & ASSD  & SO    & SD     \\ \hline

DANet \cite{fu2019dual}            & 59.45 & 43.27 & 75.88 & 52.45 & 15.12 & 2.37 & 73.98 & 66.41   \\ \hline
SENet \cite{hu2018squeeze}             & 87.65 & 78.08 & 94.23 & 82.05 & 2.59  & 0.43 & 95.61 & 95.28     \\ \hline
Attn Unet \cite{oktay2018attention}      & 87.68 & 78.16 & 91.90 & 83.94 & 5.60  & 0.56 & 94.71 & 94.33     \\ \hline
Polar   \cite{liu2021polarized}          & 87.81 & 78.36 & 91.32 & \textbf{84.67 }& \textbf{2.07}  & \textbf{0.34} & 95.44 & 95.21    \\ \hline
CBAM    \cite{woo2018cbam}         & 87.82 & 78.34 & 93.80 & 82.77 & 2.35  & 0.36 & 95.89 & 95.56   \\ \hline
SKNet   \cite{li2019selective} & \textbf{88.04} & \textbf{78.71} &\textbf{94.71}  & 82.30 & 2.70  & 0.44 &\textbf{96.24}  & \textbf{95.90}   \\ \hline

\end{tabular}}
\label{tab.22}
\end{table}

The proposed attention-based tooth segmentation framework outperforms several current 3D medical segmentation methods including DenseVoxelNet, 3D HighResNet, 3D Unet and VNet. As shown in Table \ref{tab.111}, Our proposed method has more than 25 \% DSC gain, 20 \% IOU gain, 5 \% sensitivity gain and 18 \% PPV gain, respectively. One reason why the existing methods are not effective in this task is that these networks are all relatively deep and hard to train on  CTooth with large data variance. The other reason is that current methods do not include  components e.g. attention and deep supervision, or they do not allow certain configurations to extract tooth features in tiny regions. In the proposed framework, considering the teeth size is relatively small around roots, we reduce the number of the down-sampling layers to avoid roots undetected compared to the 3D Unet.


Additionally, we augment the baseline network with several state-of-the-art attention modules. In Table \ref{tab.22}, we present tooth volume segmentation results among various attention types including channel attention (SENet and SKNet), spatial attention (Attention Unet), channel and spatial attention (CBAM), and self attention (Polar and DANet). The base segmentation framework with self attention Polar achieves the best performance on 2 spatial surface distance metrics (HD and ASD). Channel attention SKNet fully exploits the automatic selection of important channels and achieve DSC 88.04 \%, IOU 78.71 \%, SEN 94.71 \%, SO 96.24 \% and SD 95.9 \% respectively. However, DANet cannot segment tooth volumes well as a large redundant quantity of matrix multiplication operations is required in DANet. 




\section{Conclusion}
In this paper, we first introduce details of the CBCT dental dataset CTooth. This is the first open-source 3D dental CT dataset with full tooth annotations. We propose an attention-based benchmark to segment tooth regions on the CTooth dataset. Due to the fine design of the residual encoder, and the proper application of attention and deep supervision, the proposed framework achieves state-of-the-art performances compared with other 3D segmentation methods. In future, we will release more multi-organisation dental data on the next version of CTooth.
\section*{Acknowledgement}
The work was supported by the  National Natural Science Foundation of China under Grant No. U20A20386.
%
%
%
\bibliographystyle{splncs04}
\bibliography{reference}
%




\end{document}